\documentclass[sigconf]{acmart}

\usepackage[absolute]{textpos}

\usepackage{booktabs} 

\usepackage{lipsum}
\usepackage{amsmath}
\usepackage{gensymb}

\setcopyright{rightsretained}

\acmConference[arXiv 2019]{Pre-print}{June 2019}{Dayton, OH, USA}
\acmYear{2019}
\copyrightyear{2019}

\acmArticle{4}
\acmPrice{15.00}

\author{Hussein S. Al-Olimat, Valerie L. Shalin, Krishnaprasad Thirunarayan and Joy Prakash Sain}
\email{{hussein,valerie,tkprasad,joy}@knoesis.org}
\affiliation{%
\institution{Kno.e.sis Center, Wright State University, Dayton, OH.}
}

\begin{document}

\begin{textblock}{15}(0.5,0.3)
\noindent\Large \textcolor{red}{\textbf{Please cite:}} Hussein S. Al-Olimat, Valerie L. Shalin, Krishnaprasad Thirunarayan, and Joy Prakash Sain. 2019. Towards Geocoding Spatial Expressions (Vision Paper). In Proceedings of the 27th ACM SIGSPATIAL International Conference on Advances in Geographic Information Systems (SIGSPATIAL '19). Association for Computing Machinery, New York, NY, USA, 75-78. DOI:\url{https://doi.org/10.1145/3347146.3359356}
\end{textblock}

\title{Towards Geocoding Spatial Expressions}

\renewcommand{\shortauthors}{HS. Al-Olimat et al.}


\begin{abstract}

Imprecise \emph{composite} location references formed using \emph{ad hoc spatial expressions} in English text makes the geocoding task challenging for both inference and evaluation. Typically such spatial expressions fill in unestablished areas with new toponyms for finer spatial referents. For example, the spatial extent of the ad hoc spatial expression "north of" or "50 minutes away from" in relation to the toponym "Dayton, OH" refers to an ambiguous, imprecise area, requiring translation from this qualitative representation to a quantitative one with precise semantics using systems such as WGS84. Here we highlight the challenges of geocoding such referents and propose a formal representation that employs background knowledge, semantic approximations and rules, and fuzzy linguistic variables. We also discuss an appropriate evaluation technique for the task that is based on human contextualized and subjective judgment.

\end{abstract}


%
%


\maketitle


\section{Introduction}


A geographic coordinate system (GCS), such as the World Geodetic System 1984 (WGS84), quantitatively describes the exact area of locations. However, natural language refers to a spatial area qualitatively posing a challenge to precision in the face of the complexity of human thought process. Human qualitative spatial referents appear as established toponyms (e.g., "\emph{Dayton}, Ohio") or \emph{ad hoc} spatial expressions (e.g., "\emph{North of} Dayton, Ohio") linked to established/atomic toponyms with relations. In English, these often employ prepositions. The mismatch between quantitative coordinate-based referents and qualitative spatial expressions prohibits their straightforward alignment. Moreover, human spatial referents are not precise or formally defined, making them ambiguous, and laden with idiosyncrasies. According to Landau and Jackendoff's \cite{landau_jackendoff_1993} spatial cognition theory "Whatever we can talk about we can also represent". Two properties of spatial expressions challenge their representation for computing systems: they are ambiguous (i.e., requiring context) and they employ fuzzy location referents (i.e., requiring approximation). Therefore, we seek a tolerant formal representation, to serve as an \emph{interlingua} between quantitative and qualitative spatial referents acknowledging Zadeh's incompatibility principle \cite{zadeh1973outline} (i.e., "the ineffectiveness of computers in dealing with human systems").

World map projects, such as OpenStreetMap, apparently solve the problem for established toponyms. Toponyms with quantitative spatial extents are represented as polygons in GeoJSON shapefiles. E.g., the city of "Seattle, WA" is associated with a polygon that covers the whole city\footnote{\url{https://www.openstreetmap.org/relation/237385}} as defined and agreed on by the OpenStreetMap community. The existence of alternative quantitative representations, including lat/long point representations, anticipate potential disagreement in practice, where humans may not employ officially established boundaries. Moreover, only established, coarse-grained toponyms (aka. atomic toponyms) link to agreed-upon quantitative representations. However, such resources are inevitably incomplete. Humans use spatial expressions to fill in the unestablished areas with new toponyms for finer spatial referents.  
For example, "\textit{downtown} Seattle" or "\textit{northern} Seattle" refer to an unestablished, \emph{ad hoc} portion of the larger, but \emph{atomically specified toponym} (i.e., "Seattle") with imprecise and fuzzy boundaries. The number of possible mappings between this \emph{ad hoc} referent and its quantitative extent, and the lack of precise and agreed upon interpretation defies both enumeration and evaluation and therein lies the complexity.

Dealing with ad hoc spatial expressions as linguistic variables \cite{zadeh1975concept} captures the inherent fuzziness of the referents reflecting day to day usage. While a lat/long point on a map, as in WGS84 has precise and distinctive semantics, adding linguistic expressions such as "north of" or "not far from" to that point makes it difficult for spatial systems to infer the exact spatial extent of the referent because the area is not precisely defined. This requires translation from a qualitative representation to an (imprecise) quantitative one that is adequate for the intended application and context.

In this paper, we identify and then operate on polygons to represent qualitative spatial referents. A linguistic spatial expression contrasts with polygons referenced using the WGS84 system in that such linguistic expressions come with fuzzy boundaries. This representation requires us to design heuristics to draw polygons and scale them automatically \emph{based on the context of referents} (e.g., the size of the referenced toponyms \cite{CLEMENTINI1997317}). The use of an ad hoc spatial expressions serves as an \emph{approximate characterization} for a descriptive phrase of a location referent.

Standard evaluation methods that rely on precise boundaries are inherently incompatible with these new kinds of unstandardized spatial referents. Precision/Recall metrics do not readily characterize the degree of error associated with referents specified with fuzzy boundaries and are missing from gazetteers. Therefore, we suggest evaluating proposed representations using a measurement scale of a higher resolution relative to a binary decision (the Likert scale), to collect human judgment and better capture the intuition behind the use of \emph{ad hoc} spatial expressions.


\section{Related Work}

Vernacular Geography (VG) concerns the vagueness of location referents (aka. imprecise regions) that are not part of a gazetteer but are part of the regional culture (e.g., a "city center", "the downtown", or "the old downtown") \cite{hollenstein2010exploring}. These referents can be as big as the Midwest United States or as small as "downtown Dayton". Disagreement, between official boundaries of the locations and what people tend to consider as the boundaries, reflects contextual factors such as population density, housing systems, land use policy, and other social, cultural, and physical artifacts \cite{hollenstein2010exploring}. However, VG techniques largely have not developed formal linguistic rules and methods to deal with the semantics of location referents. Instead, these attempts are mostly deductive methods drawing on conclusions from patterns in geotagged data and textual descriptions.

Some toponym resolution techniques that map location names to their unambiguous spatial footprint rely on spatially annotated documents to learn the spatial distributions of document words \cite{delozier2015gazetteer}. Usually, such supervised techniques deduce the spatial footprint of words to georeference documents based on lexical choice (e.g., "Y'all" suggests origins in the"southern US"). Woodruff and Plaunt \cite{woodru1994gipsy} performed syntactic operations to retain relevant portions of phrases and omit others (e.g., retaining "Delta" from the phrase "literature on the Delta"). They use a relevance weighting method to assign weights to generated polygons for referents. They also support lexical transformation to improve coverage, for example, transforming "Valleys" into "Valley" that can then be found in a gazetteer. They strip tokens from toponyms, such as "County" from "Kern County", to create what is called "evidonyms" to retrieve locations like "Kern Water Bank" which is not part of the gazetteer. The technique considers the highest stacked polygons of all georeferenced locations as the focus area of a document to deduce the semantics of spatial expressions such as "south of the Delta". Lieberman and Samet \cite{Lieberman2348381} use geographical distances and background knowledge from gazetteers to disambiguate location references based on their relative proximity in texts. However, none of these techniques deal with isolated fine-grained and \emph{ad hoc} spatial referents in expressions that include relations such as "near" or "between".

Clementini et al. \cite{CLEMENTINI1997317} analyzed qualitative distances and defined context using a three-component frame of reference (i.e., distance system, scale, and type). Crucially, the analysis shows that distance is dependent on scale. For example, the meaning of "close" in "\emph{x} is close to \emph{y}" depends on the relative position of both locations and their respective sizes. This is very useful in practice for developing heuristics to deal with spatial expressions, especially when inferring the meaning of ad hoc spatial expressions, such as "near", that relies on the contextual toponyms in the same expression (e.g., "I live in a country near Mexico" vs. "I live in a city near Columbus, OH").


Finally, in Geographic Information Retrieval (GIR) \cite{purves2018geographic}, the understanding of spatial relationships such as "near" in the example "beaches near Dayton" requires an understanding of the theme "beaches" and the spatial extent of the toponym "Dayton". Returning all beaches within a predefined distance, or returning the closest few beaches around the atomic toponym "Dayton", regardless of distance, are alternative ways of satisfying the query. Such approaches are limited by the availability of an entry for "beaches" in the geographic coordinate system.


\section{Referencing Spatial Expressions}
\label{sec:toptypesandrefs}

Reference objects (aka. anchors), such as established atomic toponyms or any geocoded spatial extents (e.g., the northern side of a city), provide an initial frame of reference for the \emph{ad hoc} spatial referents \cite{liu2009positioning}. We acknowledge that ambiguity may persist in understanding the spatial extent of atomic toponyms. However, here, we assume that atomic toponyms are correctly geocoded and disambiguated. Additionally, we assume that all atomic and ad hoc spatial referents are correctly delimited in text using techniques such as locative expression extraction \cite{liu2014automatic} and the spatial relationships are correctly annotated using techniques such as spatial role labeling \cite{kordjamshidi2010spatial}.

The focus of this paper is to infer and geocode relative/partial spatial referents with respect to one or more identified spatial extents of anchor objects. Thus, an atomic toponym, for example, is not the head of the expression, but rather a modifier on the head, e.g., "banks" or "gas stations" in "Dayton" is more about the "banks" and "gas stations" that appear in "Dayton", than "Dayton" \emph{per se}. The task then is to specify how the \emph{"Dayton"} toponym enables the interpretation of \emph{"in"} and the scope of the referents for "banks" and "gas stations". Landau and Jackendoff \cite{landau_jackendoff_1993} suggest that representing space requires the understanding of toponyms and routes (aka. places and paths) and that prepositions extend the spatial area of a toponym to related paths and regions. The spatial relation expressed in prepositional phrases would, therefore, play the defining role of spatial extents\footnote{In this work, we will not talk about the spatial extents of routes. For example, we are not yet concerned about how to represent "via" and "through".}. In general, there are two types of spatial relations expressed in prepositional phrases:
\begin{enumerate}
    \item \textbf{Binary spatial relations:} These are the spatial referents expressed in reference to a single anchor object (e.g., "near $x$", "away\_from $x$", "beside $x$", or "in $x$").
    \item \textbf{N-ary spatial relations:} These are the spatial referents expressed in reference to two or more anchor objects (e.g., "near $x$ and $y$", "between $x$ and $y$", or "among $x$, $y$, and $z$").
\end{enumerate}

There are three main frames of reference that affect the interpretation of spatial extents and influence the semantics of spatial relations \cite{purves2018geographic}:
\begin{enumerate}
    \item \textbf{Relative:} These are the ones that are in direct relationship with the anchor objects (e.g., "in", "left\_of", or "near").
    \item \textbf{Absolute:} This uses an external frame of reference, such as the earth's cardinal directions (e.g., "north\_of" or "south\_of")
    \item \textbf{Intrinsic:} Different from the \textit{Relative} type, this adheres to the inherent properties of toponyms. For example, in "The clock tower in front of the palace", the orientation here is not important because there is only one front of the palace regardless of the relative binary relationship between the palace and the clock tower \cite{levinson2003space}.
\end{enumerate}

Other linguistic facts and cognitive constraints of language inform the understanding of spatial relations and prepositional phrases. In English, the encoding of subjects and objects of spatial prepositions reflects the size of objects \cite{landau_jackendoff_1993}. The larger sized objects/locations always appear as the object in the case of "inside", "in", "outside", etc (e.g., for "Our office is in Ohio", the area of the "office" is much smaller than the area of "Ohio"). Similarly for adjacency relationships where bigger sized spatial extents, usually of toponyms that are familiar or culturally significant, are used as a reference to smaller spatial extents, as in the example of "behind $x$". The use of "among" is interpreted as n-ary in contrast to the binary relation "between". Additionally, distances can be combined with direction markers to yield finer distinctions of spatial referents (e.g., "$5km$ away from $x$").



\section{Spatial referents representation}


Given a sentence $W= \langle w_1, w_2, \dots, w_n \rangle$, let $W$ contain one or more atomic toponyms in addition to \emph{ad hoc} spatial expressions defining the resolution of the referents and the extent of the focal area with spatial relations (see Section \ref{sec:toptypesandrefs}). While the semantics of the spatial footprint of an atomic toponym is clear (i.e., available in mapping systems and gazetteers), the spatial semantics and extent of an \emph{ad hoc} expression is not clear or formally defined. Inspired by Zadeh's work on linguistic variables and values \cite{zadeh1975concept}, we define spatial linguistic terms in ad hoc spatial expressions, such as "between" or "north\_of", by defining \emph{fuzzy} restrictions on their spatial extent.


 \begin{figure}
    \vspace{-0.3cm}
    \includegraphics[width=.85\columnwidth]{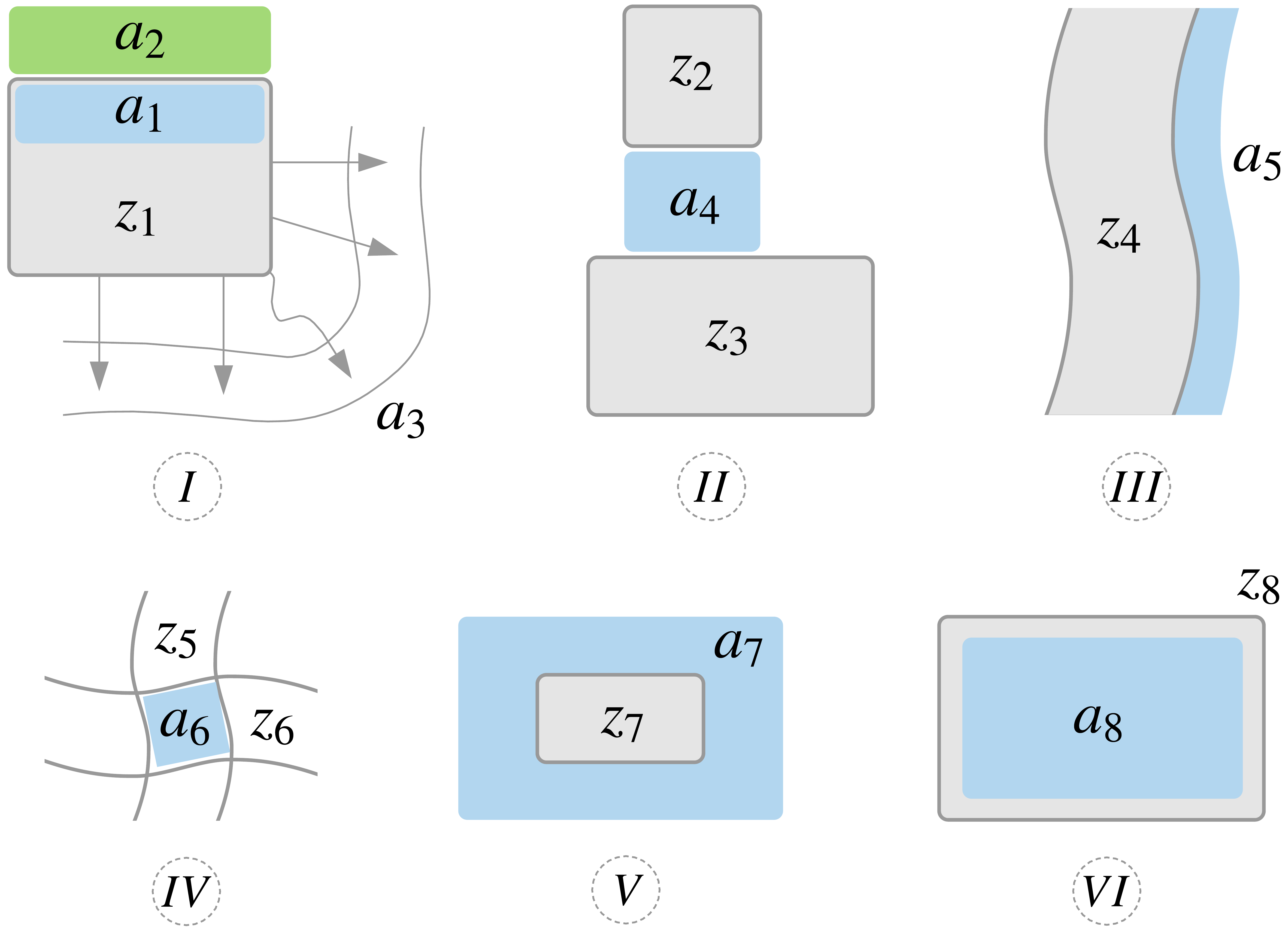}
    \centering
    \vspace{-0.3cm}
    \caption{Examples of abstract spatial representations of anchor objects ($z_{(\cdot)}$) and the ad hoc referents "northern $z_1$" ($\mathbf{a_1}$), "north\_of $z_1$" ($\mathbf{a_2}$), "$t$ time/distance away\_from $z_1$" ($\mathbf{a_3}$), "between $z_2$ and $z_3$" ($\mathbf{a_4}$), "along/on $z_4$" ($\mathbf{a_5}$), "$z5$ with $z_6$" ($\mathbf{a_6}$), "outside/out\_of $z_7$" ($\mathbf{a_7}$), and "in/inside $z_8$" ($\mathbf{a_8}$).}
    \label{fig:rep1}
    \vspace{-0.3cm}
\end{figure}


Ad hoc expressions on a linguistic value (e.g., "north\_of") such as "$t$ minutes" or "$m$ miles", modify (or make precise) the semantics induced by predefined syntactic rules. To represent qualitative spatial expressions, we expand the formulation of Zadeh's linguistic variable. We characterize a linguistic variable by a sextuple $(\mathcal{V},\mathcal{I}, \mathcal{C}, \mathcal{P}, \mathcal{G}, \mathcal{S})$, where $\mathcal{V}$ is a linguistic variable (e.g., "distance"), $\mathcal{I}$ is the set of linguistic terms for a fuzzy value or functional relation (e.g., "away\_from" and "minutes\_from"), $\mathcal{C}$ is the set of all anchor objects, $\mathcal{P}$ is the set of spatial extents of the objects in $\mathcal{C}$, $\mathcal{G}$ is the set of spatial expression syntactic rules that generates the terms in $\mathcal{I}$, and $\mathcal{S}$ is the set of semantic rules restricting the values of each term in $\mathcal{I}$ referenced using $\mathcal{C}$.

 We employ the cognitive constraints of the language from Section \ref{sec:toptypesandrefs} while constructing semantic rules to further constrain the spatial extent of referents. For example, in the case of "between", the sides of the fuzzy polygon should not exceed the edge of the shortest side of the two facing polygons as we discuss next.

Figure \ref{fig:rep1} shows abstractions of different spatial referents and the corresponding spatial expressions that can be formulated using the predefined sextuple. Below is an example formulation of $II$:


 \vspace{-10px}

\begin{align*}
& \mathcal{V} = \text{"topological"} \\
& \mathcal{I} = \text{\{ "between"; "betwixt" \}} \\
& \mathcal{C} = \{ c_1 , c_2\} \\
& \mathcal{P} = \{c_1 \to [lat_1, long_1, lat_2, long_2], \\
& \ \ \ \ \ \ \ \ \ \ c_2 \to [lat_1, long_1, lat_2, long_2] \mid c_1, c_2 \in \mathcal{C} \} \\
& \mathcal{G} = \{\langle i, c_1, \text{"and"}, c_2 \rangle \mid i \in \mathcal{I} \wedge c_1, c_2 \in \mathcal{C}\} \\
& \mathcal{S} = [\mathcal{P}(c_1).lat_2-\alpha, \mathcal{P}(c_1).long_1-\beta, \\
& \ \ \ \ \ \ \ \ \ \mathcal{P}(c_2).lat_1+\alpha, \mathcal{P}(c_1).long_2+\beta]
\end{align*}

 \noindent where $c_i$ is the string name of an anchor object (e.g., "Walmart"), $\mathcal{P}$ containing the polygon of each $c_i$ (e.g., the simple bounding box [38.40, -84.82, 42.32, -80.51]). Without loss of generality and for the convenience of illustration and notational simplicity, we narrow down the spatial representation of each anchor object from complex polygons to rectangular form (i.e., bounding boxes) with bounds on both sides (i.e., top-right $[lat_1, long_1]$ and bottom-left $[lat_2, long_2]$ points). $\mathcal{G}$ contains only one spatial expression which describes how the spatial relationship in $\mathcal{I}$ must be used (e.g., "between Walmart and Sam's Club"). Finally, the semantics of the lexical item "between" is defined as a polygon between these toponyms\footnote{For more complex polygons of the two toponyms, the area inside the convex hull would make the "between" area \cite{billen2004model}.}. As shown in Figure \ref{fig:rep1}-$II$, the area takes the shortest facing edge of the two polygons (shown as bounding boxes for ease of representation) with the height to be the distance between them. Therefore, the possible area of "between" can be $\mathcal{S}$, where $lat_{(\cdot)}$ and $long_{(\cdot)}$ are the latitudes and longitudes of the $i^{th}$ anchor object, and $\alpha$ and $\beta$ are two offsets defined based on the semantics of $I$ to create buffer distances \cite{chen2018georeferencing}.\footnote{The offsets values would be different for "between" and "on" since "on" is for exterior referents with contact while "between" does not mean contact.}

 \begin{figure}
    \vspace{-0.3cm}
    \includegraphics[width=1\columnwidth]{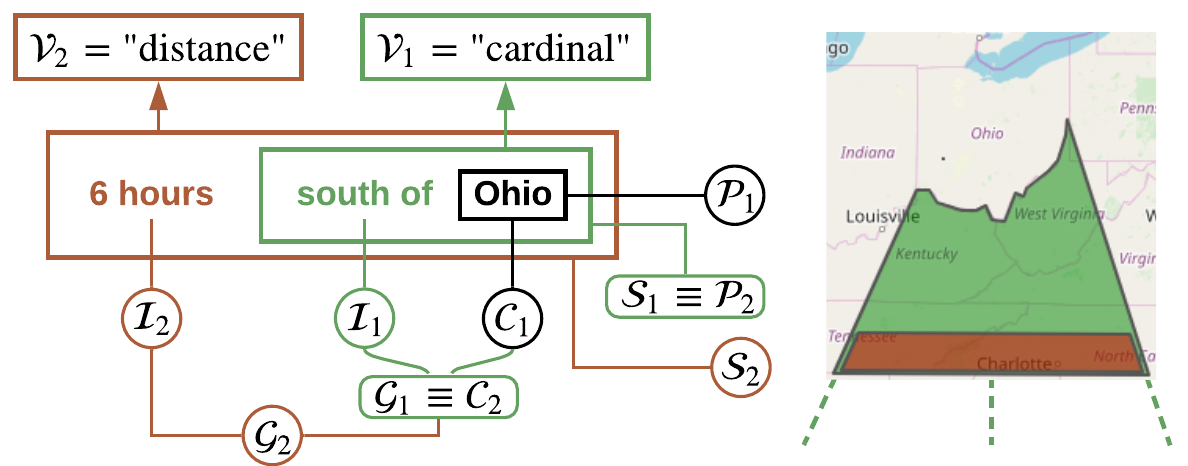}
    \centering
    \vspace{-0.5cm}
    \caption{Spatial representation of "6 hours" and "south of" ad hoc terms in relation to the atomic toponym "Ohio".}
    \label{fig:egsouthof}
    \vspace{-0.3cm}
\end{figure}

As for the directional markers in $\langle d, \text{"of"}, z \rangle \mid d \in \{ \text{"north"; "east"}; \\ \dots\}$, we create a polygon representing that area in relation to the anchor object $z$ using a cone-based cardinal direction model \cite{liu2005internal}. For example, for "south of Ohio", we create a polygon $\mathcal{S}_1$ that sits on the bottom side of Ohio's polygon $\mathcal{P}_1$ (see Figure \ref{fig:egsouthof}). The extent of $\mathcal{S}_1$ continues to a maximum of half of the equatorial circumference. Now, the constraint "6 hours" on the "south of Ohio" expression would create a sub-polygon $\mathcal{S}_2$ of the previously induced spatial extent $\mathcal{P}_2$ to only be the red polygon. Notice that since "6 hours" is fuzzy, the created polygon would capture that fuzziness by adding buffer distances to the sides of $\mathcal{S}_2$ equal to, for example, 5\% of the modifier value (i.e., $\pm18$ minutes of the $6$ hours constraint). Our formulation scales automatically with the addition of more constraints and ad hoc expressions. For example, if we add "near Asheville" to the end of "6 hours south of Ohio", we would create another polygon as in Figure \ref{fig:egsouthof}, showing the constraint on $\mathcal{S}_2$, in this case.

Similarly, we represent the examples in Figure \ref{fig:rep1}, without loss of generality, using a complete polygon (as in $\mathbf{a_1}, \mathbf{a_2}, \mathbf{a_4}, \mathbf{a_6},$ and $\mathbf{a_8}$) or a polygon with a hole in it (as in $\mathbf{a_3}$ and $\mathbf{a_7}$). 
The challenge is not only the determination of the fuzzy boundaries for spatial extents but also the range of possible interpretations of the relative position of the polygons. In the case of $\mathbf{a_3}$, there are an infinite number of points that are $t$ minutes/miles away from $\mathbf{z_1}$, which requires more context to draw the fuzzy polygon\footnote{Different distance calculation methods can be used including "as the crow flies" distance or route-based distance calculation which requires route navigation assuming "cars" as the culturally determined mode of transportation.}. Alternatively, absolute frames of reference using lexical markers such as "north\_of" or "south\_of" constrain the area of the referent, but without context or other lexical markers, we would draw a polygon with a hole in it covering all possible spatial extents around the atomic toponym. Finally, the alternative intrinsic frame of reference needs a full understanding of the surrounding area and requires a gazetteer with rich metadata and toponym semantics. For example, finding the front of a university campus might require searching for clusters of points of interest to suggest the orientation of the toponym. We leave further exploration and evaluation of this as future work.


\section{Evaluation}

The majority of the techniques in the literature focus on geocoding the spatial extent of atomic toponyms (i.e., Toponym Resolution) \cite{Kamalloo2018}. The default evaluation methods are Precision, Recall, and the F1 measure as the task is to link a toponym to a gazetteer record or compare the geocoordinates with annotated toponyms in documents. Other measures include micro and macro averaging, mean error distance, AUC of geocoding errors, and Acc@161, which calculates the accuracy of geocoding within a 161km tolerance distance \cite{gritta2018pragmatic}. Clearly, such evaluation methods are incompatible with the qualitative, imprecise location referents that have no records in a gazetteer, have fuzzy meaning, and are finer than typical atomic toponyms with (sometimes) a huge error tolerance of 161km.

As there is no satisfactory approach in the literature or gold data available for fine-grained referents involving imprecise spatial regions, we devise an approach using a Likert scale, similar to \cite{aflaki2018challenges} but while adding the negative side of the scale (i.e., including "Strongly Agree" to "Strongly Disagree"). Effectively, the individual scores ranging from 1 to -1 to be plugged in the following equation to score all geocodings:

\begin{equation}
    \nonumber
    Score=\frac{1}{mn}\sum_{i=0}^{n} \sum_{j=0}^{m} \sigma_{i, j}
\end{equation}

\noindent
where $n$ is the number of spatial expressions, $m$ is the number of annotators, $\sigma$ is the weight (1 to -1 Likert scale) given to a geocoded expression by the annotator $j$, and $Score$ is the cumulative system score averaged over all weights capturing the judges agreements and disagreements to evaluate the geocoder.


\section{Conclusions and Future Work}

In this paper, we devised a formal representation to geocode imprecise and \emph{ad hoc} location referents as opposed to atomic gazetteer toponyms. The potential disagreement between the different interpretations and the unavailability of appropriate gold standard data made conventional evaluation methods incompatible with the task at hand and required from us to develop more approximate evaluation measures. This paper constitutes an initial step in this complex task, and we plan to continue this work by expanding on the formulation and the semantic rules, and evaluating our solution on real-world data.




\bibliographystyle{ACM-Reference-Format}
\bibliography{refs}

\end{document}